\documentclass{article}
\usepackage{graphicx}
\usepackage{xcolor}
\usepackage{soul}
\usepackage[ruled,vlined]{algorithm2e}
\usepackage{amsmath}
\usepackage{gensymb}
\usepackage{bbm}
\usepackage{bm}
\usepackage[final]{corl_2020} 
\usepackage{amsfonts}
\usepackage{amsmath}
\usepackage{lipsum,multicol}
\usepackage{wrapfig,lipsum,booktabs}
\title{Deep Reinforcement Learning with Population-Coded Spiking Neural Network for Continuous Control}

\author{
  Guangzhi Tang, Neelesh Kumar, Raymond Yoo, and Konstantinos P. Michmizos\\
  Department of Computer Science\\
  Rutgers University\\
  \texttt{\{gt235, nk525, rby6, michmizos\}@cs.rutgers.edu} \\
}

\begin{document}
\maketitle


\begin{abstract}
The energy-efficient control of mobile robots has become crucial as the complexity of their real-world applications increasingly involves high-dimensional observation and action spaces, which cannot be offset by their limited on-board resources. An emerging non-Von Neumann model of intelligence, where spiking neural networks (SNNs) are executed on neuromorphic processors, is now considered as an energy-efficient and robust alternative to the state-of-the-art real-time robotic controllers for low dimensional control tasks. The challenge now for this new computing paradigm is to scale so that it can keep up with real-world applications. To do so, SNNs need to overcome the inherent limitations of their training, namely the limited ability of their spiking neurons to represent information and the lack of effective learning algorithms. Here, we propose a population-coded spiking actor network (PopSAN) that was trained in conjunction with a deep critic network using deep reinforcement learning (DRL). The population coding scheme, which is prevalent across brain networks, dramatically increased the representation capacity of the network and the hybrid learning combined the training advantages of deep networks with the energy-efficient inference of spiking networks. To show that our approach can be used for general-purpose spike-based reinforcement learning, we demonstrated its integration with a wide spectrum of policy-gradient based DRL methods covering both on-policy and off-policy DRL algorithms. We deployed the trained PopSAN on Intel's Loihi neuromorphic chip and benchmarked our method against the mainstream DRL algorithms for continuous control. To allow for a fair comparison among all methods, we validated them on OpenAI gym tasks. Our Loihi-run PopSAN consumed 140 times less energy per inference when compared against the deep actor network on Jetson TX2, and achieved the same level of performance. Our results demonstrate the overall efficiency of neuromorphic controllers and suggest the hybrid reinforcement learning approach as an alternative to deep learning, when both energy-efficiency and robustness are important.
\end{abstract}



\section{Introduction}
Mobile robots with continuous high-dimensional observation and action spaces are increasingly being deployed to solve complex real-world tasks. Given their limited on-board energy resources, there is an unmet need to design energy-efficient solutions for the continuous control of these autonomous robots. Deep reinforcement learning (DRL) methods based on policy-gradient have been successful in learning optimal control policies for complex tasks~\citep{ha2018automated, zhu2017target}; However, their optimality comes at the cost of high energy consumption, rendering them ill-suited for several applications~\citep{dong2017more}.

An energy-efficient alternative to deep networks is provided by spiking neural networks (SNNs) deployed on neuromorphic processors. In this emerging neuromorphic computing paradigm, where memory and computation are tightly integrated, neurons perform asynchronous, event-based computations~\citep{davies2018loihi}. Mounting studies are suggesting SNNs as low-energy solutions for several real-world robotic problems~\citep{tang2019spiking, taunyazov20event, michaelis2020robust}. For robotic control, SNN approaches are typically based on reward-modulated local learning rules~\cite{bing2018end, fremaux2013reinforcement} that perform well in low-dimensional tasks but often fail in complex problems, where optimization becomes difficult in the absence of a global loss function~\citep{legenstein2005can}. Recently,~\citep{rosenfeld2019learning} proposed a policy gradient-based algorithm to train an SNN for learning stochastic policies. However, the algorithm operates over a discrete action space, with a rather limited use on high-dimensional continuous control problems.  

To address the limitations of SNN in solving high-dimensional continuous control problems, one approach is to combine the energy-efficiency of SNN with the optimality of DRL. To this end, a popular SNN construction method is to directly convert a trained deep neural network (DNN) into an SNN using weight and threshold balancing~\citep{patel2019improved}. The main problems with this approach are that it often results to a spiking network with an inferior performance to the corresponding DNN, and also requires large timesteps for inference that dramatically increases the energy cost~\citep{rathi2020enabling}. To overcome this, a recent work proposed a hybrid learning algorithm where an SNN with rate-coded inputs is trained using DRL to learn optimal control policies for mapless navigation of a mobile robot in a static environment~\citep{tang2020reinforcement}. However, this method suffers in complex high-dimensional tasks where the optimality of the control policy highly depends on the encoding precision of individual spiking neurons that have limited representation ability~\citep{averbeck2006neural}. The practicality of this solution becomes even less when a small inference timestep is used for higher energy-efficiency, since this is expected to further reduce the representation ability of the neurons as they encode data using their firing rate.

Interestingly, abstracting away the brain's topology and its computational principles has recently given rise to the design of SNNs that exhibit human-like behavior~\citep{balachandar2020spiking} and improved performance~\cite{kreiser2020chip}. A key attribute in the brain associated with efficient computation is the use of populations of neurons to represent information, from sensory stimuli to output signals, where each neuron in a population has a receptive field that captures part of the encoded signal~\citep{georgopoulos1986neuronal}. Indeed, initial studies on the population coding scheme have demonstrated its ability to better represent the stimuli~\citep{tkavcik2010optimal}, which led to recent successes in training SNNs for complex high-dimensional supervised learning tasks~\citep{bellec2018long, pan2019neural}. The demonstrated effectiveness of the population coding opens up prospects for developing efficient population-coded SNNs that can learn optimal solutions for high-dimensional continuous control tasks.

In this paper, we propose a population-coded spiking actor network (PopSAN) that is trained using DRL algorithms to learn energy-efficient solutions for continuous control problems\footnote{Code available at https://github.com/combra-lab/pop-spiking-deep-rl}. At the core of our PopSAN lies its ability to encode each dimension of the observation and action spaces in individual neuron populations with learnable receptive fields, effectively increasing the representation capacity of the network. Since different control tasks require specialized DRL solutions~\citep{henderson2018deep}, we integrated our PopSAN with both on-policy and off-policy DRL algorithms, in particular, DDPG~\citep{lillicrap2015continuous}, TD3~\citep{fujimoto2018addressing}, SAC~\citep{haarnoja2018soft}, and PPO~\citep{schulman2017proximal}, thereby demonstrating its applicability to a wide spectrum of policy-gradient based DRL algorithms. We deployed the trained PopSAN on Intel's Loihi neuromorphic processor and evaluated our method on OpenAI gym tasks with rich and unstable dynamics that are used in benchmarking continuous control algorithms. We compared our method on its rewards gained and energy consumption against the mainstream DRL algorithms. Our Loihi-run PopSAN consumed 140 times less energy per inference when compared against the deep actor network on Jetson TX2, while also achieving the same level of performance. These results introduce the DRL algorithms to the spiking domain, scaling them as neuromorphic solutions to reinforcement learning tasks where energy efficiency matters.


\section{Methods}
\label{sec:methods}
\subsection{Population-coded Spiking Actor Network (PopSAN) embedded into DRL algorithms}
\begin{figure}
    \centering
    \includegraphics{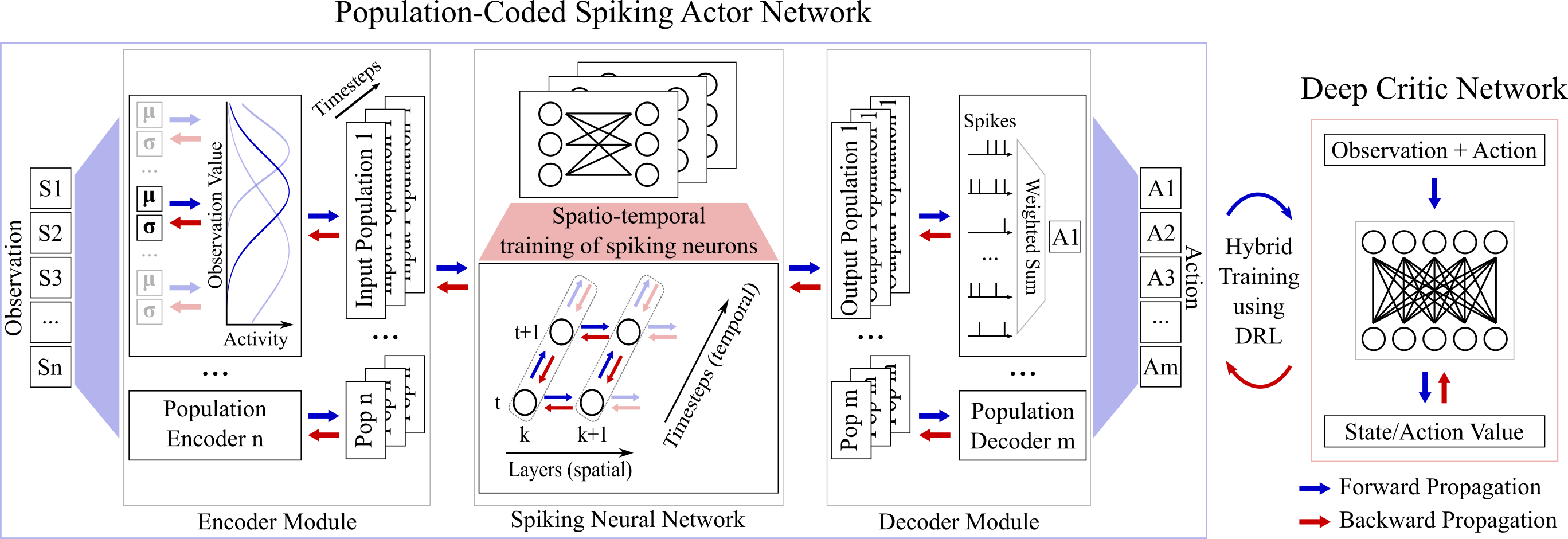}
    \caption{Population-coded spiking actor network (PopSAN) was trained in conjunction with a deep critic network using the DRL algorithms. Neurons in the input populations encoded each observation dimension and drove a multi-layered and fully connected SNN. At the end of forward timesteps, the activities of each output population was decoded into its corresponding action dimension. }
    \label{fig:fig1}
\end{figure}

We propose a population-coded spiking actor network (PopSAN) that is trained in conjunction with a deep critic network using the DRL algorithms. During training, the PopSAN generated an action $a \in \mathbb{R}^N$ for a given observation, $s$, and the deep critic network predicted the associated state-value $V(s)$ or action-value $Q(s,a)$, which in turn optimized the PopSAN, in accordance with a chosen DRL method (Fig.~\ref{fig:fig1}). The encoder module in the PopSAN encoded each dimension of the observation into the activity of an individual neuron population. During forward propagation, the input populations drove a multi-layered and fully-connected SNN to produce activities of output populations which were then decoded into their corresponding action dimensions at the end of every $T$ timesteps (Algorithm~\ref{alg:forward}).

To build the SNN, we used the current-based leaky-integrate-and-fire (LIF) model of a spiking neuron. The dynamics of the LIF neurons are governed by a 2 step model as described in Algorithm \ref{alg:forward}: i) integrating the presynaptic spikes $\mathbf{o}$ into current $\mathbf{c}$; and ii) integrating the current $\mathbf{c}$ into membrane voltage $\mathbf{v}$; $d_c$ and $d_v$ are the current and voltage decay factors. Subsequently, the neuron fires a spike if its membrane potential exceeds a threshold. We used the hard-reset model where the membrane potential is reset to rest potential upon spiking. The resultant spikes are transmitted to the post-synaptic neurons at the same inference timestep, assuming zero propagation delay.

\begin{algorithm}
\SetAlgoLined

Randomly initialize weight matrices $\textbf{W}$ and biases $\textbf{b}$ for each SNN layer;\\

Initialize encoding means \bm{$\mu$} and standard deviations \bm{$\sigma$} for all input populations;\\

Randomly initialize decoding weight vectors $\textbf{W}_d$ and bias $b_d$ for each action dimension;\\

$N$-dimensional observation, $\textbf{s}$;\\

Spikes from input populations generated by the encoder module : \textbf{X} = Encoder($\textbf{s}$, \bm{$\mu$}, \bm{$\sigma$});\\

\For{t=1,...,T}{
Spikes from input populations at timestep t: $\textbf{o}^{(t)(0)}=\textbf{X}^{(t)}$\;
    \For{k=1,...,K}{
    Update LIF neurons in layer $k$ at timestep $t$ based on spikes from layer $k-1$:\\
    $\textbf{c}^{(t)(k)} = d_c \cdot \textbf{c}^{(t-1)(k)}+ \textbf{W}^{(k)}\textbf{o}^{(t)(k-1)} + \textbf{b}^{(k)}$\;
    $\textbf{v}^{(t)(k)} = d_v \cdot \textbf{v}^{(t-1)(k)}\cdot (1-\textbf{o}^{(t-1)(k)}) + \textbf{c}^{(t)(k)}$\;
    $\textbf{o}^{(t)(k)} = Threshold(\textbf{v}^{(t)(k)})$\;
    }
}
$M$-dimensional action $\textbf{a}$ generated by the decoder module:\\
Sum up the spikes of output populations: $\textbf{sc} = \sum_{t=1}^{T}\textbf{o}^{(t)(K)}$\;
\For{i=1,...,M}{
Compute firing rates of the $i^{th}$ output population : $\textbf{fr}^{(i)} = \textbf{sc}^{(i)} /\ T$;\\
Compute $i^{th}$ dimension of action: $a^{i} = \textbf{W}_d^{(i)} \cdot \textbf{fr}^{(i)} + b_d^{(i)}$;
}
 \caption{Forward propagation through PopSAN}
 \label{alg:forward}
\end{algorithm}

Our PopSAN is functionally equivalent to a deep actor network and can be integrated with any actor-critic based DRL algorithm. Specifically, we integrated the PopSAN with both on-policy and off-policy DRL algorithms, namely DDPG~\citep{lillicrap2015continuous}, TD3~\citep{fujimoto2018addressing}, SAC~\citep{haarnoja2018soft}, and PPO~\citep{schulman2017proximal}, as follows: For DDPG and TD3, we trained the PopSAN to predict the action for which the trained critic network generated the maximum action-value. For SAC, we trained the PopSAN to predict the mean of the stochastic action distribution, and a deep network to predict its standard deviation. This was done by minimizing the distance between the probability of the action sampled from the predicted distribution and the predicted action-value generated by the trained critic network. For PPO, the PopSAN was trained to predict the mean of the action distribution by optimizing the clipped surrogate loss.

\subsection{Population encoding and decoding in PopSAN}
We encoded each dimension of the observation and action space into the activities of individual input and output population of spiking neurons. The encoder module converted the continuous observation into spikes in the input populations, and the decoder module decoded the output population activities into real-valued actions.

For the $i^{th}$ dimension of the $N$-dimensional observation, $s_i$, $i \in \{1...N\}$, we created a population of neurons, $E_i$, to encode it. Dropping the $i$ for notational simplicity, the neurons in $E$ had Gaussian receptive fields (\bm{$\mu$}, \bm{$\sigma$}). The \bm{$\mu$} were initialized to be evenly distributed in the space of $s$ and \bm{$\sigma$} were preset to be large enough to ensure non-zero population activity in the entire space of $s$.

The encoder module computed the activity of the population $E$ in two phases: It first transformed the observation values into the stimulation strength for each neuron in the population, $\mathbf{A_E}$:
\begin{equation}
\mathbf{A_E} = EXP(-1/2 \cdot ((s - \bm{\mu}) / \bm{\sigma})^2) \label{eq:3}
\end{equation}
Second, the computed $\mathbf{A_E}$ was used to generate the spikes of the neurons in $E$. There are two possible ways to do this: i) Probabilistic encoding, where spikes for all the neurons were generated at each timestep with the probabilities defined by $\mathbf{A_E}$; and ii) Deterministic encoding, where the neurons in $E$ were simulated as one-step soft-reset IF neurons, with $\mathbf{A_E}$ acting as the presynaptic inputs to the neurons. The dynamics of the neurons were governed by the following equation:
\begin{equation}
\begin{gathered}
\mathbf{v}(t) = \mathbf{v}(t-1) + \mathbf{A_E} \\
 o_k(t) = 1\textrm{ \&  } v_k(t) = v_k(t) - (1 - \epsilon), \quad  \textrm{if } v_k(t) > 1 - \epsilon 
    \label{eq:4}
\end{gathered}
\end{equation}
where $k$ denotes the index of a neuron in $E$ and $\epsilon$ is a small constant. For both types of encoders, \bm{$\mu$} and \bm{$\sigma$} are task-specific trainable parameters. In our experiments, we employed both types of encoders for the second phase. 

The output layer of the SNN comprised of populations of neurons, where a population $D_i$ represented dimension $i$ of the $M$-dimensional action, $a_i$,  $i \in \{1...M\}$. Dropping the $i$ for notational simplicity, the decoder module decoded the activity of the output population, $D$, into its corresponding real-valued action in two phases: First, after every $T$ timesteps, the spikes of each neuron in $D$ were summed up to obtain the firing rate $\textbf{fr}$ over $T$. Second, the action $a$ was returned as the weighted sum of the computed $\textbf{fr}$ (Algorithm~\ref{alg:forward}). The receptive fields of the output populations were formed by their connection weights which were learned as part of the training. 

\subsection{PopSAN training}
We used gradient descent to update the PopSAN parameters where the exact loss function depended upon the chosen DRL algorithm, as explained in Section 2.1.  
The gradient of the loss with respect to the computed action $\nabla _{\mathbf{a}}L$ was used to train the parameters of PopSAN. 

The parameters for each output population $i$, $i \in {1,...,M}$ were updated independently as follows:

\begin{equation}
    \nabla _{\mathbf{W}_d^{(i)}}L = \nabla _{a_i}L \cdot \mathbf{W}_d^{(i)} \cdot \mathbf{fr}^{(i)} \quad, \quad \nabla _{b_d^{(i)}}L = \nabla _{a_i}L \cdot \mathbf{W}_d^{(i)}
\end{equation}

The SNN parameters were updated using the extended spatiotemporal backpropagation introduced in~\citep{tang2020reinforcement}. We used the rectangular function $\mathnormal{z(v)}$, defined in~\citep{wu2018spatio}, to approximate the gradient of a spike. The gradient of the loss with respect to the SNN parameters for each layer $k$ were computed by collecting the gradients backpropagated from all the timesteps:
\begin{equation}
    \nabla _{\textbf{W}^{(k)}} L = \sum_{t=1}^{T}\textbf{o}^{(t)(k-1)}\cdot\nabla _{\textbf{c}^{(t)(k)}} L \quad, \quad
    \nabla _{\textbf{b}^{(k)}} L = \sum_{t=1}^{T}\nabla _{\textbf{c}^{(t)(k)}} L \label{eq:9}
\end{equation}

Lastly, we updated the parameters independently for each input population $i$, $i \in {1,...,N}$ as follows:
\begin{equation}
    \nabla_\mathbf{\bm{\mu}^{(i)}}L = \sum_{t=1}^{T}\nabla_{\mathbf{o}^{(t)(0)}_i}L \cdot \mathbf{A_E}^{(i)} \cdot \frac{s_i - \bm{\mu}^{(i)}}{\bm{\sigma}^{(i)^2}}\ ,\ \nabla_\mathbf{\bm{\sigma}^{(i)}}L = \sum_{t=1}^{T}\nabla_{\mathbf{o}^{(t)(0)}_i}L \cdot \mathbf{A_E}^{(i)} \cdot \frac{(s_i - \bm{\mu}^{(i)})^2}{\bm{\sigma}^{(i)^3}}
\end{equation}

We updated all parameters every $T$ timesteps. For a step-by-step analysis of the flow of gradients during the training, we direct the readers to Section 1 of the supplementary material. 

\begin{figure}
    \centering
    \includegraphics{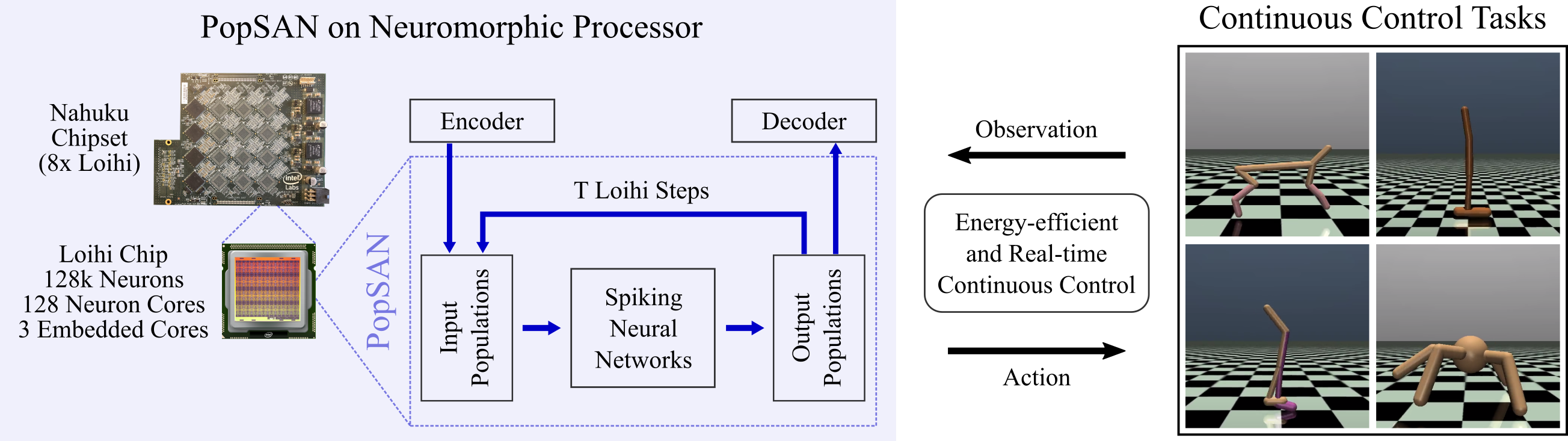}
    \caption{PopSAN deployed on Intel's Loihi neuromorphic processor for energy-efficient and real-time continuous control. Loihi interfaced with the continuous control task environment in real-time using the interaction framework. Blue arrows indicate the sequence of operations inside the Loihi chip.}
    \label{fig:fig2}
\end{figure}

\subsection{Energy-efficient continuous control with Intel's Loihi neuromorphic chip}
We deployed the trained PopSAN on Intel's Loihi neuromorphic chip (Fig.~\ref{fig:fig2}). To this end, we introduced an interaction framework that enabled Loihi to control the agents in the OpenAI gym in real-time. To reduce the communication overhead, the first phase of the encoding (computation of $\mathbf{A_E}$) was carried out on the computer that hosted the task environment, and the second phase (spike generation) was performed on the low-frequency x86 chip embedded on Loihi. Likewise, the first phase of the decoding (computation of $\mathbf{fr}$) was performed on the embedded chip, and the second phase (action computation) was performed on the host computer.

We then used the layer-wise rescaling technique to map the trained PopSAN with full-precision weights onto the low-precision loihi chip~\citep{tang2020reinforcement}. Lastly, we forced each layer in the SNN on Loihi to start its operation one timestep after its previous layer started its operation. This was done because the postsynaptic neurons on Loihi receive the presynaptic spikes in the next timestep of their operation, as opposed to GPUs where they are received at the same timestep. 
\section{Experiments and Results}
\label{sec:results}
The goals of our experiments were the following: i) To demonstrate the integration of PopSAN with both on-policy and off-policy DRL algorithms by benchmarking the performance of our method against the corresponding deep actor networks; ii) To demonstrate the need for population coding through comparison with the state-of-the-art SNN approaches and examining the effect of learning in neuron populations; and iii) To demonstrate PopSAN's advantage in performing energy-efficient and real-time continuous control when deployed on Loihi. We evaluated our method on the OpenAI gym~\citep{1606.01540} tasks with rich and unstable dynamics that are commonly used for benchmarking continuous control algorithms. To limit the effect of initialization, we trained ten models for each algorithm corresponding to ten random seeds. Each model was trained for 1 million steps and evaluated every 10k steps by testing it using the deterministic policy outputted by the actor. To compensate for the effect of randomness in the tasks, we computed the average rewards over 10 episodes for each evaluation, where each episode lasted for a maximum of 1000 execution steps.

\begin{figure}
    \centering
    \includegraphics{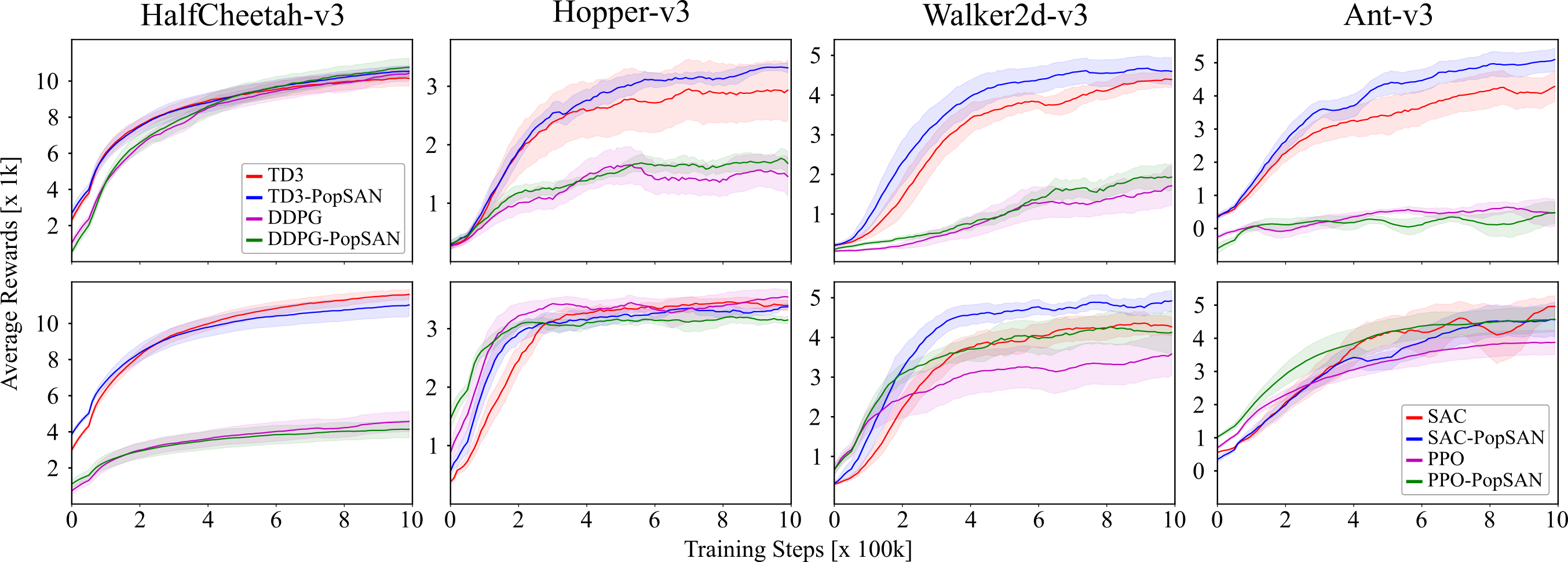}
    \caption{PopSAN trained using on-policy and off-policy DRL algorithms achieved the same level of performance as the deep actor networks across all the environments. Figure shows the mean rewards and half the value of standard deviation (top panel: TD3, DDPG; bottom panel: SAC, PPO). Plots are smoothed for clarity.}
    \label{fig:fig3}
\end{figure}

\subsection{Benchmarking PopSAN against mainstream DRL algorithms}
We compared the performance of PopSAN trained using the on-policy and off-policy DRL algorithms against the corresponding deep actor networks. Our method achieved the same level of performance as the deep actor networks across all the tasks for all the DRL algorithms (Fig.~\ref{fig:fig3}), indicating that the two approaches are
functionally equivalent. 
The hyperparameters for training can be found in Section 2 of the supplementary material. 

\subsection{Benchmarking PopSAN against other SNN design approaches}
We compared our method against the following two recently suggested approaches for integrating SNN with DRL: i) DNN to SNN conversion method (DNN-SNN) in which a deep actor network is trained using the DRL algorithms and is converted to an SNN using weight rescaling~\citep{patel2019improved}. In the converted SNNs, the post-spike membrane potential of the LIF neurons can be set to rest potential (hard-reset; H) or a positive potential to retain information about the previous spikes (soft-reset; S) which demonstrated better performance in a recent study~\citep{han2020rmp}. For implementation details, we direct the readers to Section 3 of the supplementary material. ii) SAN with rate-coded inputs (RateSAN) that uses single neuron representation to encode the inputs and the outputs and is trained using the hybrid learning algorithm~\citep{tang2020reinforcement}. The experiments reported here were performed using the TD3 algorithm.

\begin{figure}
    \centering
    \includegraphics{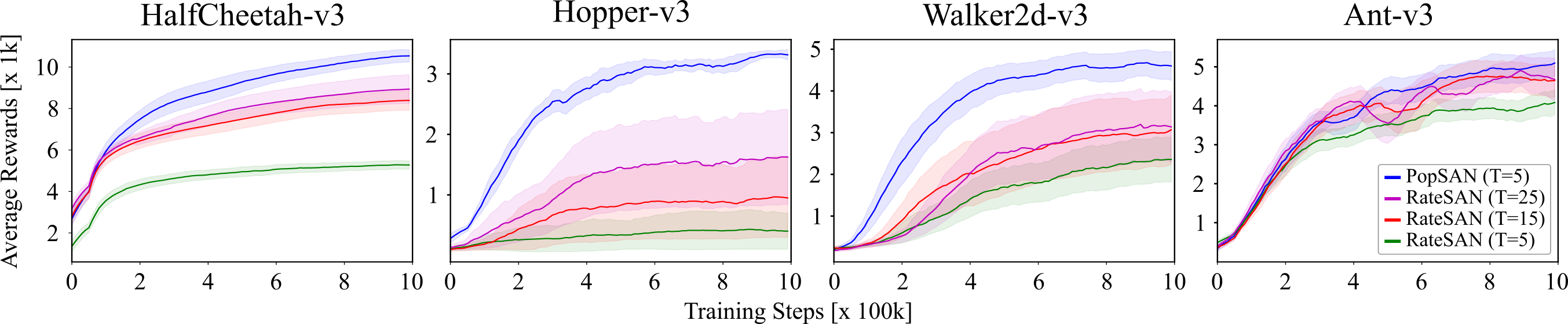}
    \caption{PopSAN (T=5) consistently performs better than RateSAN (T=5,15,25). All models were trained using TD3 algorithm.  }
    \label{fig:fig4}
\end{figure}

The two SNN approaches failed to match the performance of our method even when trained with a value of $T$ that was 5 times higher (Fig. \ref{fig:fig4}; Table \ref{tab:Loihi}). This could be because both the SNN approaches had a limited representation capacity; While the DNN-SNN method suffered from loss in precision during conversion, the RateSAN method had an inherent limitation on the representation capacity of individual neurons. This is further supported by the large performance decrease in the Hopper task which has highly unstable dynamics.

\subsection{Learning in neuron populations}

To further demonstrate the need for population coding, we evaluated the effect of the input and output neuron population size on the performance of the PopSAN trained using TD3, and investigated how learning influenced the representation capacity of the input and output neuron populations. 

\begin{figure}
    \centering
    \includegraphics{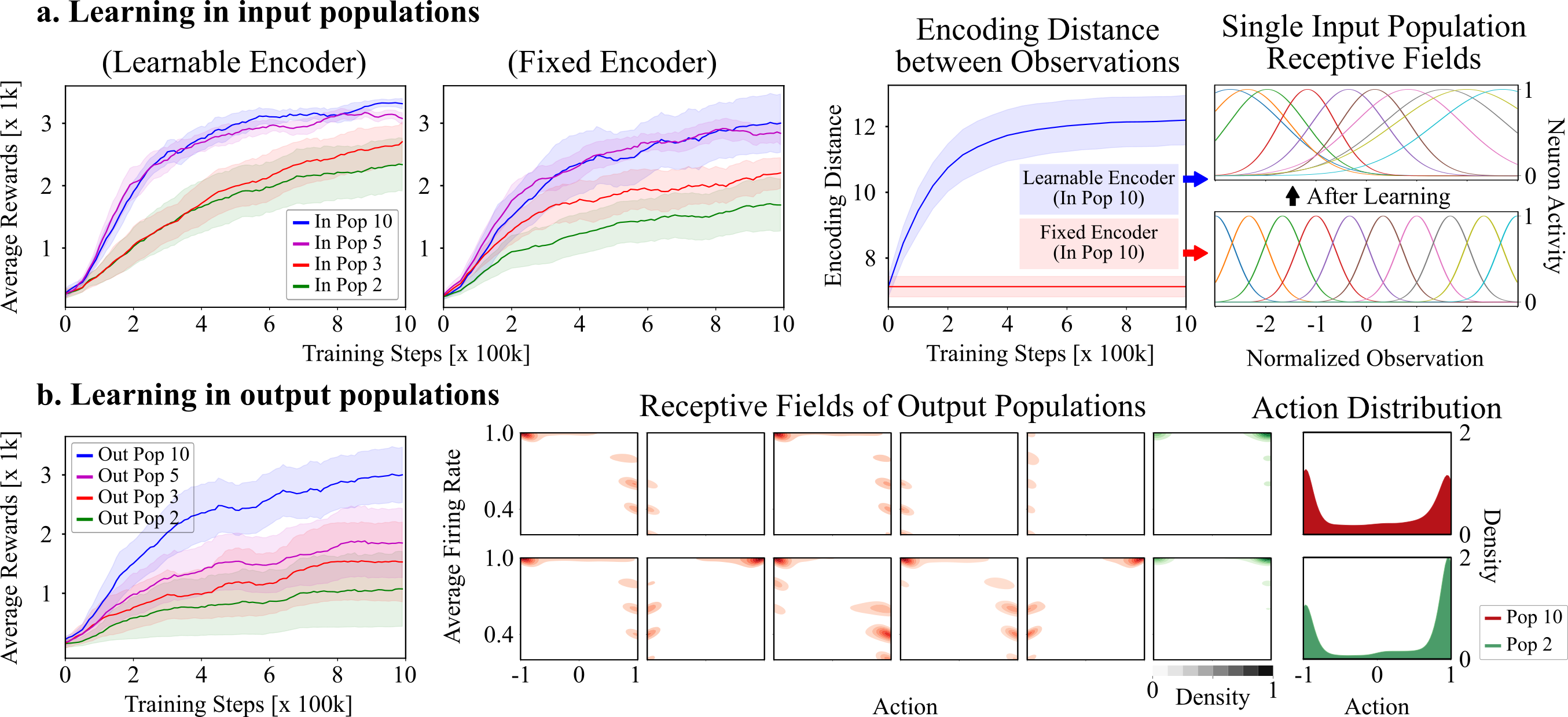}
    \caption{Learning in neuron populations for PopSAN trained using TD3 for Hopper-v3. \textbf{a.} Learning in the input populations led to a better input representation and increase in performance. \textbf{b.} Larger output populations size learned redundant and better action representations and resulted in higher rewards. Receptive fields for each neuron in the output population corresponding to an action.}
    \label{fig:fig5}
\end{figure}

First, we trained PopSAN with different input population sizes per observation dimension: 2, 3, 5, 10 while keeping the output population size fixed to 10. Fig.~\ref{fig:fig5}a shows that a decrease in the input population size hurt the performance of PopSAN. To investigate the effect of learning in the encoder, we trained PopSAN with fixed encoders in which the encoder parameters (\bm{$\mu$}, \bm{$\sigma$}) remained unchanged during training. The learnable encoder performed better than the fixed encoder (Fig. \ref{fig:fig5}a) for all input population sizes. A possible reason for the superior performance of the learnable encoder could be its ability to better separate different observations in its encoding. To justify this hypothesis, we computed the average L2 distance between the spike encodings of different observations for fixed and learnable encoders. The learnable encoder resulted in an encoding that increased the distance between the different observations (Fig. \ref{fig:fig5}a), thereby suggesting that it learned better input representations. 

Next, we trained PopSAN with different output population sizes per action dimension: 2, 3, 5, 10 while keeping the input population size fixed to 10. The performance improved with the increasing output population size (Fig. \ref{fig:fig5}b). To inspect the learned action representations of the PopSAN, we computed the receptive fields of the output population neurons by estimating the joint probability density of the neuron activity and the predicted action values using kernel density estimation. Fig. \ref{fig:fig5}b shows that PopSAN with larger population size learned redundant representations of action and could cover a wider range of action values. This suggests that PopSAN with large output population size can learn better action representation. 

\subsection{Evaluating continuous control on Loihi}
To validate PopSAN's applicability in performing energy-efficient and real-time continuous control, we deployed it on Loihi. Our approach on Loihi exhibited high performance in terms of rewards gained (Table \ref{tab:Loihi}) with only a marginal decrease when compared to PopSAN on a full-precision GPU. We then computed the inference speed and energy consumption for the continuous control of HalfCheetah-v3 using i) deep actor network on CPU (E5-1600), GPU (Tesla K40), embedded AI chip (Jetson TX2- energy-efficient (Q) and high-performance (N) modes), and  ii) PopSAN on Loihi (Nahuku 8). We computed the energy cost per inference as the ratio of the dynamic power and the number of inferences performed per second. Our PopSAN on Loihi was 140 times more energy-efficient than the deep actor network on the low-power processor for DNNs- Jetson TX2 (Table \ref{tab:hardware}), and also had high inference speed, which enabled real-time control.

\begin{table}
    \caption{Max average rewards over 10 models for PopSAN and DNN-SNN.}
    \centering
    \begin{tabular}{c c c c c}
    \hline
    (T, Reset, Device) & HalfCheetah & Hopper & Walker2d & Ant\\
    \hline
    PopSAN (5, H, GPU) & \textbf{10729} ($\sigma$-644) & \textbf{3599} ($\sigma$-68) & \textbf{4995} ($\sigma$-682) & \textbf{5416} ($\sigma$-673)\\
    PopSAN (5, H, Loihi) & 10505 ($\sigma$-636) & 3289 ($\sigma$-292) & 4280 ($\sigma$-987) & 5220 ($\sigma$-625)\\
    DNN-SNN (5, H, GPU) & 3991 ($\sigma$-925) & 996 ($\sigma$-683) & 391 ($\sigma$-359) & -35 ($\sigma$-743)\\
    DNN-SNN (5, S, GPU) & 5256 ($\sigma$-931) & 1129 ($\sigma$-879) & 768 ($\sigma$-786) & -7 ($\sigma$-719)\\
    DNN-SNN (15, H, GPU) & 6989 ($\sigma$-937) & 1704 ($\sigma$-1215) & 1804 ($\sigma$-1286) & 148 ($\sigma$-793)\\
    DNN-SNN (15, S, GPU) & 9729 ($\sigma$-859) & 2385 ($\sigma$-1403) & 4116 ($\sigma$-724) & 1634 ($\sigma$-1116)\\
    DNN-SNN (25, H, GPU) & 7913 ($\sigma$-955) & 1932 ($\sigma$-1393) & 2475 ($\sigma$-1318) & 347 ($\sigma$-664)\\
    DNN-SNN (25, S, GPU) & 10722 ($\sigma$-806) & 2523 ($\sigma$-1346) & 4565 ($\sigma$-1033) & 3026 ($\sigma$-1718)\\
    \hline
    \end{tabular}
    \label{tab:Loihi}
\end{table}


\section{Discussion and Conclusion}
\label{sec:conclusion}
In this paper, we presented PopSAN, a population-coded SNN, that achieved the same level of performance as deep networks while being two orders of magnitudes more energy-efficient. The similarity in performance was consistent when we integrated PopSAN with both on-policy and off-policy DRL methods. This demonstrates its applicability to a wide spectrum of policy-gradient based DRL algorithms, highlighting its use in solving several reinforcement learning tasks including continuous control.

The population coding scheme increased the representation capacity of the network and led to better encodings of observations and actions. This enabled the PopSAN trained in conjunction with a deep network to effectively learn complex high-dimensional continuous control tasks with less timesteps for inference. While most prior works on SNNs have a predominant focus on the network design and training~\citep{bellec2018long, wu2018spatio}, our work demonstrates that an appropriate learnable encoding of the network inputs and outputs with the right inductive priors can lead to a substantial increase in the performance.

The PopSAN running on Loihi was 140 times more energy-efficient on continuous control tasks than the deep actor network running on TX2, an embedded power-efficient AI chip. This result becomes particularly important in real-world mobile robot applications such as disaster-relief and planetary exploration where on board resources are limited. Such applications typically rely on multimodal sensing for robustness~\citep{liu2017learning} and require high-dimensional actions for dexterity~\citep{ha2018automated}. DRL methods overcome the problem of high-dimensionality of the inputs and outputs through the use of large networks such as a deep convolutional neural network~\citep{zhu2017target}. The use of such large networks, however, significantly increases the energy costs for control~\citep{dong2017more}. On the other hand, our proposed method can potentially decrease the energy costs by a large factor for such applications, while further improvements in energy-efficiency are expected by employing event-based sensors~\cite{maqueda2018event} and memristive neuromorphic processors~\citep{ankit2017resparc}, which are much more efficient than their digital counterparts.

We showed here that the PopSAN can be successfully integrated with both on-policy and off-policy DRL algorithms, highlighting its applicability to a wide variety of tasks that require specialized DRL solutions~\citep{henderson2018deep}. This is particularly significant in real-world reinforcement learning applications where several factors such as sample efficiency of the algorithm, stochasticity of the environment, objectives of the reward function, and safety constraints determine which DRL algorithm would be most appropriate for a given task~\citep{dulac2019challenges}. Overall, our proposed spike-based solution can become a strong alternative to deep learning for real-world reinforcement learning applications, when both energy-efficiency and robustness are important.

\begin{table}
    \caption{Power performance and inference speed across hardware}
    \centering
    \begin{tabular}{c c c c c c}
    \hline
    Method & Device & Idle (W) & Dynamic (W) & Inf/s & $\mu$J/Inf\\
    \hline
    DNN & CPU & 15.51 & 58.93 & 7450 & 7909.86\\
    DNN & GPU & 24.68 & 46.04 & 3782 & 12174.46\\
    DNN & TX2(N) & 1.24 & 1.76 & 750 & 2346.71\\
    DNN & TX2(Q) & 1.05 & 0.73 & 438 & 1670.94\\
    PopSAN & Loihi & 1.084 & 0.003 & 226 & \textbf{11.98}\\
    \hline
    \end{tabular}
    \label{tab:hardware}
\end{table}



\clearpage
\acknowledgments{This work is supported by Intel's NRC Grant Award.}


\bibliography{example}  

\clearpage
\begin{center}
\textbf{\large Supplementary Materials: Deep Reinforcement Learning with Population-Coded Spiking Neural Network for Continuous Control}
\end{center}

\setcounter{section}{0}
\setcounter{figure}{0}
\setcounter{footnote}{0}
\renewcommand{\thefigure}{S\arabic{figure}}

\renewcommand*{\theHsection}{chX.\the\value{section}}

\section{PopSAN training using backpropagation}

Here, we analyze the step-by-step flow of the gradients during the training of PopSAN. The gradient of the loss with respect to the computed action,  $\nabla _{\mathbf{a}}L$ was used to train the parameters of PopSAN. 

The parameters for each output population $i$, $i \in {1,...,M}$ were updated independently as follows:

\begin{equation}
\begin{gathered}
\nabla _{\mathbf{fr}^{(i)}}L = \nabla _{a_i}L \cdot \mathbf{W}_d^{(i)}\\
    \nabla _{\mathbf{W}_d^{(i)}}L = \nabla _{\mathbf{fr}^{(i)}}L \cdot \mathbf{fr}^{(i)} \quad, \quad \nabla _{b_d^{(i)}}L = \nabla _{\mathbf{fr}^{(i)}}L
\end{gathered}
\end{equation}

The SNN parameters were updated using the extended spatiotemporal backpropagation for which we used the rectangular function equation \eqref{eq:pseudo} to approximate the gradient of a spike.
\begin{equation}
      \mathnormal{z(v)} =
  \begin{cases}
  1 & \text{if $|v - V_{th}| < a$} \\
  0 & \text{otherwise}
  \end{cases}
   \label{eq:pseudo}
  \end{equation}
where $z$ is the pseudo-gradient, $a$ is the threshold window for passing the gradient.

For each timestep, $t < T$, the flow of the gradients through the SNN can be described as follows:

At the output population layer $K$, we have:
\begin{equation}
    \nabla _{\textbf{sc}} L = \frac{1}{T}\cdot\nabla _{\textbf{fr}} L \quad, \quad
     \nabla _{\textbf{o}^{(t)(K)}} L = \nabla _{\textbf{sc}} L \label{eq:6}
\end{equation}
Then for each layer, $k = K$ down to $1$: 
\begin{equation}
\begin{gathered}
    \nabla _{\textbf{v}^{(t)(k)}} L = z(\textbf{v}^{(t)(k)})\cdot\nabla _{\textbf{o}^{(t)(k)}} L + d_v(1-\textbf{o}^{(t)(k)})\cdot\nabla _{\textbf{v}^{(t+1)(k)}} L 
\\
    \nabla _{\textbf{c}^{(t)(k)}} L = \nabla _{\textbf{v}^{(t+1)(k)}} L + 
    d_c\nabla _{\textbf{c}^{(t+1)(k)}} L  \\
    \nabla _{\textbf{o}^{(t)(k-1)}} L = \textbf{W}^{(k)'}\cdot \nabla _{\textbf{c}^{(t)(k)}} L \label{eq:7}
\end{gathered}
\end{equation}
When $t=T$, the gradients with respect to voltage and current do not have the second additive term of \eqref{eq:7} since the temporal gradients backpropagated from the future timesteps are absent.
The gradient of the loss with respect to the SNN parameters for each layer $k$ can then be computed by collecting the gradients backpropagated from all the timesteps:
\begin{equation}
    \nabla _{\textbf{W}^{(k)}} L = \sum_{t=1}^{T}\textbf{o}^{(t)(k-1)}\cdot\nabla _{\textbf{c}^{(t)(k)}} L \quad, \quad
    \nabla _{\textbf{b}^{(k)}} L = \sum_{t=1}^{T}\nabla _{\textbf{c}^{(t)(k)}} L \label{eq:9}
\end{equation}

Lastly, we computed the gradient of the loss with respect to the parameters for each input population $i$, $i \in {1,...,N}$. For simplicity, we directly backpropagated the gradient w.r.t. the neuron stimulation strength $\mathbf{A_E}$ regardless of whether a spike was generated or not at any timestep t.

\begin{equation}
\begin{gathered}
\nabla_{\mathbf{A_E}^{(i)}}\mathbf{o}^{(t)(1)}_i=1\quad, \quad\nabla_{\mathbf{A_E}^{(i)}}L = \sum_{t=1}^{T}\nabla_{\mathbf{o}^{(t)(0)}_i}L\\
    \nabla_\mathbf{\bm{\mu}^{(i)}}L = \nabla_{\mathbf{A_E}^{(i)}}L \cdot \mathbf{A_E}^{(i)} \cdot \frac{s_i - \bm{\mu}^{(i)}}{\bm{\sigma}^{(i)^2}}\quad, \quad \nabla_\mathbf{\bm{\sigma}^{(i)}}L = \nabla_{\mathbf{A_E}^{(i)}}L \cdot \mathbf{A_E}^{(i)} \cdot \frac{(s_i - \bm{\mu}^{(i)})^2}{\bm{\sigma}^{(i)^3}}
\end{gathered}
\end{equation}

We updated all the parameters of PopSAN after every $T$ timesteps.

\section{Hyperparameters for training regular DRL and PopSAN}

Here, we describe the implementation details and hyperparameter configurations for the deep actor network and PopSAN. Our experiments were built upon the open-source codebases from OpenAI Spinning Up \footnote{https://github.com/openai/spinningup}, OpenAI Baselines\footnote{https://github.com/openai/baselines}, and Stable Baselines\footnote{https://github.com/hill-a/stable-baselines}. PopSAN training used the same hyperparameters as the deep actor network unless explicitly stated. Hyperparameter configurations for all methods were as follows:

\begin{itemize}
    \item \textbf{DDPG (Deep actor network)}
    \begin{itemize}
        \item[$-$] Actor network (256, relu, 256, relu, tanh); Critic network (256, relu, 256, relu, linear)
        \item[$-$] Actor learning rate $1e-3$; Critic learning rate $1e-3$
        \item[$-$] Reward discount factor $\gamma=0.99$
        \item[$-$] Gaussian exploration noise with stddev 0.1
        \item[$-$] Maximum length of replay buffer $1e6$
        \item[$-$] Soft target update factor $0.005$
        \item[$-$] Batch size 100
    \end{itemize}
    \item \textbf{TD3 (Deep actor network)}
    \begin{itemize}
        \item[$-$] Actor network (256, relu, 256, relu, tanh); Critic network (256, relu, 256, relu, linear)
        \item[$-$] Actor learning rate $1e-3$; Critic learning rate $1e-3$
        \item[$-$] Reward discount factor $\gamma=0.99$
        \item[$-$] Gaussian exploration noise with stddev 0.1
        \item[$-$] Gaussian smoothing noise for target policy with stddev 0.2
        \item[$-$] Maximum length of replay buffer $1e6$
        \item[$-$] Soft target update factor $0.005$
        \item[$-$] Batch size 100
    \end{itemize}
    \item \textbf{SAC (Deep actor network)}
    \begin{itemize}
         \item[$-$] Actor network (256, relu, 256, relu, (mean: tanh, log\_stddev: linear), Gaussian)
        \item[$-$] Critic network (256, relu, 256, relu, linear)
        \item[$-$] Actor learning rate $1e-3$; Critic learning rate $1e-3$
        \item[$-$] Reward discount factor $\gamma=0.99$
        \item[$-$] Entropy regularization coefficient $\alpha=0.2$
        \item[$-$] Maximum length of replay buffer $1e6$
        \item[$-$] Soft target update factor $0.005$
        \item[$-$] Batch size 100
    \end{itemize}
    \item \textbf{PPO (Deep actor network)}
    \begin{itemize}
        \item[$-$] Actor network (256, relu, 256, relu, tanh) + policy log\_stddev variable
        \item[$-$] Critic network (256, relu, 256, relu, linear)
        \item[$-$] Actor + log\_stddev variable learning rate $1e-4$; Critic learning rate $1e-4$
        \item[$-$] Reward discount factor $\gamma=0.99$; GAE $\lambda=0.95$
        \item[$-$] Clip ratio 0.2; Entropy coefficient 0.001
        \item[$-$] Optimize epochs per iteration 25
        \item[$-$] Number of parallel environments 10
        \item[$-$] Maximum length of replay buffer $1e6$
        \item[$-$] Critic loss discount factor $0.5$
        \item[$-$] Batch size 100
    \end{itemize}
    \item \textbf{DDPG, TD3 (PopSAN)}
    \begin{itemize}
        \item[$-$] PopSAN (In Pop, 256, LIF, 256, LIF, Out Pop)
        \item[$-$] Population size for single observation and action dimension, 10
        \item[$-$] PopSAN learning rate $1e-4$
        \item[$-$] HalfCheetah \& Ant: Deterministic encoding
        \item[$-$] Hopper \& Walker2d: Probabilistic encoding
    \end{itemize}
    \item \textbf{SAC (PopSAN)}
    \begin{itemize}
        \item[$-$] PopSAN for policy mean (In Pop, 256, LIF, 256, LIF, Out Pop)
        \item[$-$] Network for policy log\_stddev (256, relu, 256, relu, linear)
        \item[$-$] Population size for single observation and action dimension, 10
        \item[$-$] PopSAN learning rate $1e-4$; log\_stddev network learning rate $1e-3$
        \item[$-$] Deterministic encoding for all tasks
    \end{itemize}
    \item \textbf{PPO (PopSAN)}
    \begin{itemize}
        \item[$-$] PopSAN for policy mean (In Pop, 256, LIF, 256, LIF, Out Pop)
        \item[$-$] Population size for single observation and action dimension, 10
        \item[$-$] PopSAN + log\_stddev learning rate $5e-6$
        \item[$-$] Deterministic encoding for all tasks
    \end{itemize}
\end{itemize}

\begin{figure}[b]
    \centering
    \includegraphics{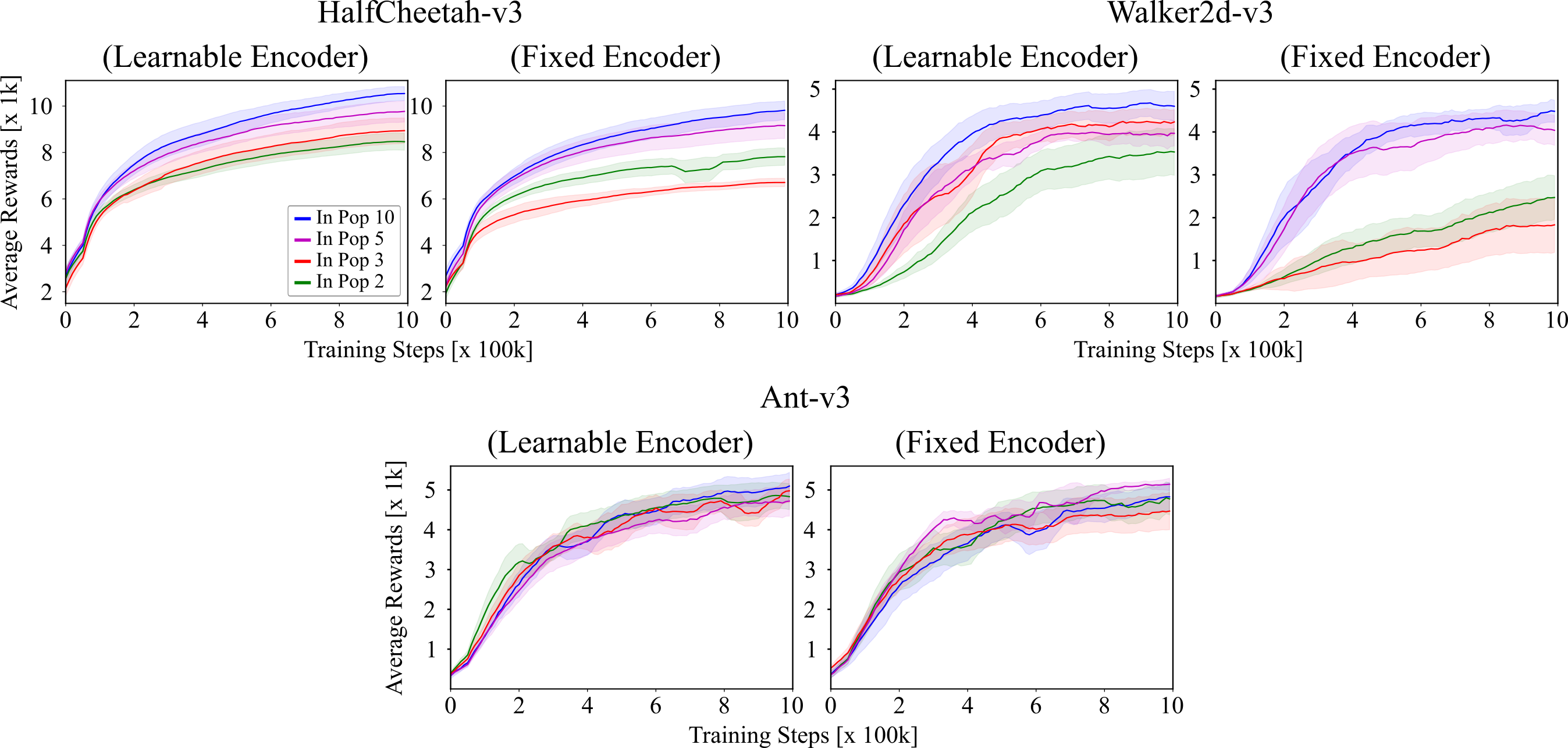}
    \caption{Learning in the input populations led to a better input representation and an increase in the performance across all environments.}
    \label{fig:sfig1}
\end{figure}

\begin{figure}
    \centering
    \includegraphics{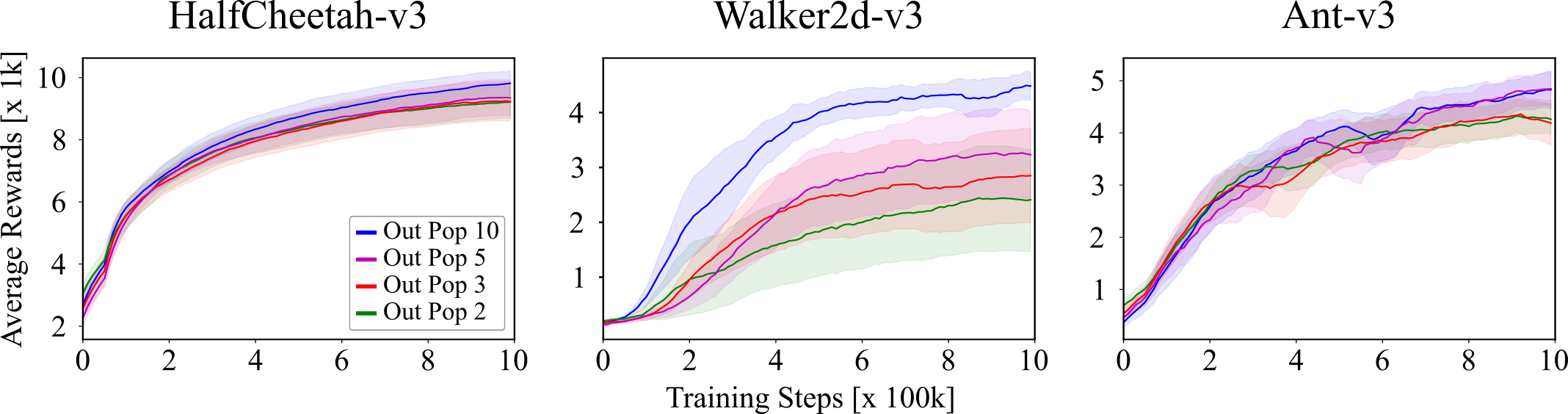}
    \caption{Larger output populations size resulted in higher rewards for all environments.}
    \label{fig:sfig2}
\end{figure}

\section{DNN to SNN conversion method}

The DNN to SNN conversion (DNN-SNN) method converted a trained DNN to SNN using weight rescaling and grid searching as follows: first, a deep actor network was trained using a chosen DRL algorithm. To overcome the limited representation of the converted SNN and for a fair comparison, the DNN had the same architecture as the PopSAN. Second, the parameters from the trained DNN were directly used for the SNN with appropriate rescale factors. Since the firing rate of spiking neurons is limited to the range [0,1], the maximum activation of each DNN layer needed to be rescaled to unity to allow the SNN to represent the full range of DNN activations. To do this, the DNN-SNN method set the bias-rescale factor for each layer to be equal to the maximum output of the current layer computed during training and the weight-rescale factor to be the ratio of the maximum output of the previous layer and the current layer. Third, to improve the performance of the converted SNN, the network weights and biases were further rescaled by a factor between 0.1-1.0 (determined using grid search) on randomly sampled episodes. We chose the factor with the highest average reward over 10 episodes of evaluation.

\section{Power measurement details}

We measured the average power consumed and the speed of performing inference for the observations recorded during testing. The power consumption and inference speed were computed by taking the average of the measurements of 10 runs with each run comprising of inferences over 100k observations recorded during testing. For power measurement, we used software tools that probed the on-board sensors of each device: powerstat for CPU, nvidia-smi for GPU, sysfs for TX2, and energy probe for Loihi. To accurately measure the power consumption of PopSAN on Loihi, we deployed 8 networks on the Nahuku chipset at the same time. The power consumption was then computed by averaging over the number of networks deployed.

\section{Additional ablation studies for neuron populations}
Figures \ref{fig:sfig1} and \ref{fig:sfig2} show the results for PopSAN trained with different input and output population sizes: 2, 3, 5, 10 for all other environments not covered in the main text. 

\end{document}